%% file: bionetics.tex
\documentclass[conference]{IEEEtran}

\usepackage[
pdftitle={Digital Ecosystems: Evolving Service-Oriented Architectures}, 
pdfauthor={Gerard Briscoe},
pdfsubject={Ecosystem-Oriented Architectures},
colorlinks=true, linkcolor=black, citecolor=black,  filecolor=black, urlcolor=black,
hyperfootnotes=false]{hyperref}

\usepackage{graphicx, setspace, acronym, ifthen, substr, color, nnfootnote}

\usepackage[paper=a4paper, asymmetric, left=1.5cm,right=1.5cm,top=2cm,bottom=2cm]{geometry}

\newcommand{\execute}[1]{\immediate\write18{#1}}

\newboolean{final}\setboolean{final}{true}

\newcommand{\Circ}[1]{\textcircled{\scriptsize #1}}

\newcommand{\tfigure}[9]
	{
	\IfSubStringInString{!}{#7}{\begin{figure}[#7]}{\begin{figure}[!t]}
	\IfSubStringInString{mm}{#8}{\vspace{#8}}{}
	\centering
	
	\IfSubStringInString{pdf}{#3}
		{
		\execute{cd images; ln -s #2.pdf .#2.gdf}
		\includegraphics[#1]{images/#2}
		}
		{\IfSubStringInString{graph}{#3}
			{
			\includegraphics[#1]{images/#2}
			}
			{
			\includegraphics[#1]{images/#2-crop.pdf}
			}
		}
		
	\vspace{#6}
	\caption[#4]
		{
		\label{#2}
		\tcaption{#4}{#5}
		}
	\IfSubStringInString{mm}{#9}{\vspace{#9}}{}
	\end{figure}
	}

\newcommand{\italics}{\textit}

\newcommand{\tcaption}[2]
	{
	\IfSubStringInString{:}{#2}{\italics{#1 #2}}{\italics{#1: #2}}
	}

\def\blfootnote{\xdef\@thefnmark{}\@footnotetext}

\newcommand{\setCap}[2]{#1\immediate\write18{./mkcaption.sh #2}}
\newcommand{\getCap}[1]{\acl{#1}}

\begin{document}

\input{acronyms}

\title{Digital Ecosystems:\\Evolving Service-Oriented Architectures\pdfbookmark[1]{Title}{Title}} 

\author{\authorblockN{Gerard Briscoe}
\authorblockA{Intelligent Systems and Networks Group\\
Department of Electrical and Electronic Engineering\\
Imperial College London\\
London, United Kingdom, SW7 2BT\\ 
e-mail: gerard.briscoe@ic.ac.uk}
\and
\authorblockN{Philippe De Wilde}
\authorblockA{Intelligent Systems Lab\\
School of Mathematical and Computer Science\\
Heriot-Watt University\\
Edinburgh, United Kingdom, EH14 4AS\\
e-mail: pdw@macs.hw.ac.uk}}
\maketitle

\begin{abstract}
\pdfbookmark[1]{Abstract}{Abstract}
We view Digital Ecosystems to be the \emph{digital counterparts of biological ecosystems}, exploiting the self-organising properties of biological ecosystems, which are considered to be robust, self-organising and scalable architectures that can automatically solve complex, dynamic problems. Digital Ecosystems are a novel optimisation technique where the optimisation works at two levels: a first optimisation, migration of agents (representing services) which are distributed in a decentralised peer-to-peer network, operating continuously in time; this process feeds a second optimisation based on evolutionary computing that operates locally on single peers and is aimed at finding solutions to satisfy locally relevant constraints. We created an \ac{EOA} of Digital Ecosystems by extending \acp{SOA} with \ac{DEC}, allowing services to recombine and evolve over time, constantly seeking to improve their effectiveness for the user base. Individuals within our Digital Ecosystem will be applications (groups of services), created in response to user requests by using evolutionary optimisation to aggregate the services. These individuals will migrate through the Digital Ecosystem and adapt to find \emph{niches} where they are useful in fulfilling other user requests for applications. Simulation results imply that the Digital Ecosystem performs better at large scales than a comparable \ac{SOA}, suggesting that incorporating ideas from theoretical ecology can contribute to useful self-organising properties in digital ecosystems.
\end{abstract}

\IEEEpeerreviewmaketitle

\input{captions}

\nnfoottext{This work is supported by the European Commission under the EU project Digital Business Ecosystems (contract number 507953 \cite{DBE}).}

\section{Introduction}
Is mimicking ecosystems the future of information systems? A key challenge in modern computing is to develop systems that address complex, dynamic problems in a scalable and efficient way, because the increasing complexity of software makes designing and maintaining efficient and flexible systems a growing challenge \cite{newsArticle1, slashdot, newsArticle3}. What with the ever expanding number of services being offered online from \acp{API} being made public, there is an ever growing number of computational units available to be combined in the creation of applications. However, this is currently a task done manually by programmers, and it has been argued that current software development techniques have hit a \emph{complexity wall} \cite{lyytinen2001nwn}, which can only be overcome by automating the search for new algorithms. There are several existing efforts aimed at achieving this automated service composition \cite{reef3, reef5, reef1, reef4}, the most prevalent of which is \aclp{SOA} and its associated standards and technologies \cite{curbera2002uws, SOAstandards}. 

Alternatively, nature has been in the research business for 3.8 billion years and in that time has accumulated close to 30 million \emph{well-adjusted} solutions to a plethora of design challenges that humankind struggles to address with mixed results \cite{biomimicry}. Biomimicry is a discipline that seeks solutions by emulating nature's designs and processes, and there is considerable opportunity to learn elegant solutions for human-made problems \cite{biomimicry}. Biological ecosystems are thought to be robust, scalable architectures that can automatically solve complex, dynamic problems, possessing several properties that may be useful in automated systems. These properties include self-organisation, self-management, scalability, the ability to provide complex solutions, and automated composition of these complex solutions \cite{Levin}.

Therefore, an approach to the aforementioned challenge would be to develop Digital Ecosystems, artificial systems that aim to harness the dynamics that underlie the complex and diverse adaptations of living organisms in biological ecosystems. While evolution may be well understood in computer science under the auspices of \emph{evolutionary computing} \cite{eiben2003iec}, ecological models are not. The possible connections between Digital Ecosystems and their biological counterparts are yet to be closely examined, so potential exists to create an \acl{EOA} with the essential elements of biological ecosystems, where the word \emph{ecosystem} is more than just a \emph{metaphor}. We propose that an ecosystem inspired approach, would be more effective at greater scales than traditionally inspired approaches, because it would be built upon the scalable and self-organising properties of biological ecosystems \cite{Levin}.

\section{Service-Oriented Architectures}

Our approach to evolving high-level software applications requires a modular reusable paradigm to software development. Service-oriented architectures (SOAs) are the current state-of-the-art approach, being the current iteration of interface/component-based design from the 1990s, which was itself an iteration of event-oriented design from the 1980s, and before then modular programming from the 1970s \cite{histOfProg, krafzig2004ess}. Service-oriented computing promotes assembling application components into a loosely coupled network of services, to create flexible, dynamic business processes and agile applications that span organisations and computing platforms \cite{papazoglou2003soc}. This is achieved through a \acs{SOA}, an architectural style that guides all aspects of creating and using business processes throughout their life-cycle, packaged as services. This includes defining and provisioning the infrastructure that allows different applications to exchange data and participate in business processes, loosely coupled from the operating systems and programming languages underlying the applications \cite{soa1w}. Hence, a \acs{SOA} represents a model in which functionality is decomposed into distinct units (services), which can be distributed over a network, and can be combined and reused to create business applications \cite{papazoglou2003soc}.

A \acs{SOA} depends upon service-orientation as its fundamental design principle. In a \acs{SOA} environment, independent services can be accessed without knowledge of their underlying platform implementation \cite{soa1w}. Services reflect a \emph{service-oriented} approach to programming that is based on composing applications by discovering and invoking network-available services to accomplish some task. This approach is independent of specific programming languages or operating systems, because the services communicate with each other by passing data from one service to another, or by co-ordinating an activity between two or more services \cite{papazoglou2003soc}. So, the concepts of \acsp{SOA} are often seen as built upon, and the development of, the concepts of modular programming and distributed computing \cite{krafzig2004ess}.

\acsp{SOA} allow for an information system architecture that enables the creation of applications that are built by combining loosely coupled and interoperable services \cite{soa1w}. They typically implement functionality most people would recognise as a service, such as filling out an online application for an account, or viewing an online bank statement \cite{krafzig2004ess}. Services are intrinsically un-associated units of functionality, without calls to each other embedded in them. Instead of services embedding calls to each other in their source code, protocols are defined which describe how services can talk to each other, in a process known as orchestration, to meet new or existing business system requirements \cite{singh2005soc}. This is allowing an increasing number of third-party software companies to offer software services, such that \acs{SOA} systems will come to consist of such third-party services combined with others created in-house, which has the potential to spread costs over many users and uses, and promote standardisation both in and across industries \cite{chhatpar2008}. For example, the travel industry now has a well-defined and documented, set of both services and data, sufficient to allow any competent software engineer to create travel agency software using entirely off-the-shelf software services \cite{kotok2001eng, cardoso2005isw}. Other industries, such as the finance industry, are also making significant progress in this direction \cite{zimmermann2004sgw}.

The vision of \acsp{SOA} assembling application components from a loosely coupled network of services that can create dynamic business processes and agile applications that span organisations and computing platforms, is visualised in Figure \ref{SOAvecFinal}. It will be made possible by creating compound solutions that use internal organisational software assets, including enterprise information and legacy systems, and combining these solutions with external components residing in remote networks \cite{SOApaper0}. The great promise of \acsp{SOA} is that the \emph{marginal cost} of creating the n-th application is virtually zero, as all the software required already exists to satisfy the requirements of other applications. Only their \emph{combination} and \emph{orchestration} are required to produce a new application \cite{tang2004ews, modi2008}. The \emph{key} is that the interactions between the \emph{chunks}, are not specified within the \emph{chunks} themselves. Instead, the interaction of services (all of whom are hosted by un-associated peers) is specified by users in an ad-hoc way, with the intent driven by newly emergent business requirements \cite{leymann2002wsa}. 

\tfigure{scale=1.0}{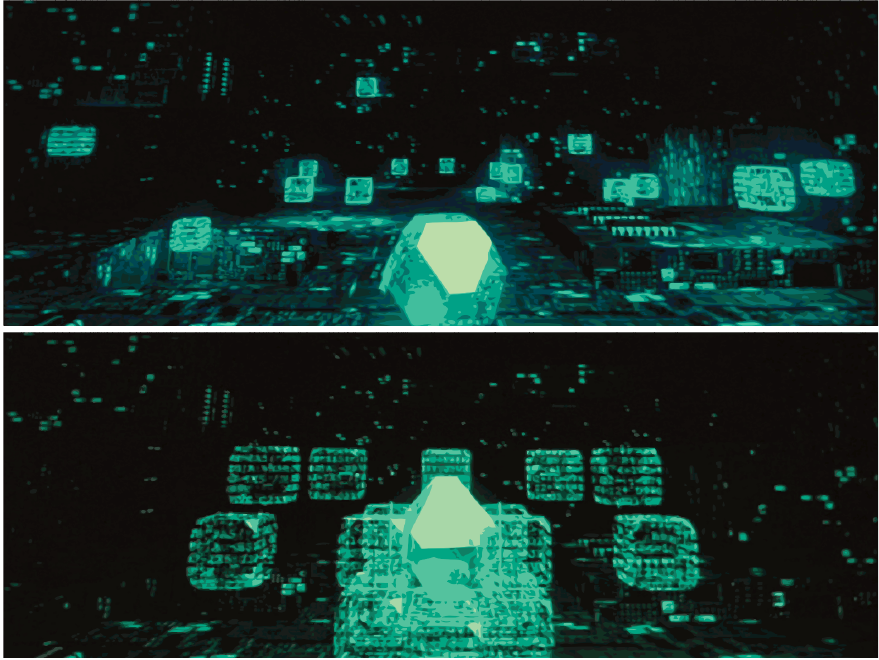}{graffle}{Service-Oriented Architectures}{Abstract visualisations, with the first image showing the loosely joined services as cuboids, and the service orchestration as a polyhedron; and the second image showing their high interoperability and re-usability in forming applications, from the use of standardised interfaces and external service orchestration.}{-8mm}{}{}{-4mm}

The pinnacle of \acs{SOA} interoperability, is the exposing of services on the internet as \emph{web services} \cite{soa1w}. A web service is a specific type of service that is identified by a \ac{URI}, whose service description and transport utilise open Internet standards. Interactions between web services typically occur as \ac{SOAP} calls carrying \ac{XML} data content. Interface descriptions of the web services are expressed using the \ac{WSDL} \cite{SOApaper2}. The \ac{UDDI} standard defines a protocol for directory services that contain web service descriptions. \ac{UDDI} enables web service clients to locate candidate services and discover their details. Service clients and service providers utilise these standards to perform the basic operations of \acsp{SOA} \cite{SOApaper2}. Service aggregators can then use the \ac{BPEL} to create new web services by defining corresponding compositions of the interfaces and internal processes of existing services \cite{SOApaper2}.

\acs{SOA} services inter-operate based on a formal definition (or contract, e.g. \ac{WSDL}) that is independent of the underlying platform and programming language. Service descriptions are used to advertise the service capabilities, interface, behaviour, and quality \cite{SOApaper2}. The publication of such information about available services provides the necessary means for discovery, selection, binding, and composition of services \cite{SOApaper2}. The (expected) behaviour of a service during its execution is described by its behavioural description (for example, as a workflow process). Also, included is a \ac{QoS} description, which publishes important functional and non-functional service quality attributes, such as service metering and cost, performance metrics (response time, for instance), security attributes, integrity (transactional), reliability, scalability, and availability \cite{SOApaper2}. Service clients (end-user organisations that use some service) and service aggregators (organisations that consolidate multiple services into a new, single service offering) utilise \emph{service descriptions} to achieve their objectives \cite{SOApaper2}. One of the most important and continuing developments in \acsp{SOA} is the use of \emph{semantic descriptions} for service discovery, so that a client can discover the services semantically, and then apply transformations to adapt the interface of the services to the interface expected, using already available client software \cite{SOAsemantic}. 

There are multiple standards available and still being developed for \acsp{SOA} \cite{SOAstandards}, most notably of recent being \ac{REST} \cite{singh2005soc}. The software industry now widely implements a thin SOAP/WSDL/UDDI veneer atop existing applications or components that implement the web services paradigm \cite{SOApaper0}, but the choice of technologies could change with time. Therefore, \acsp{SOA} and its services are best defined generically, because \acsp{SOA} are technology agnostic and need not be tied to a specific technology \cite{papazoglou2003soc}. Within the current and future scope of \acsp{SOA}, there is clearly potential to \emph{evolve} complex high-level software applications from the modular services of \acsp{SOA}, instead of the instruction level evolution currently prevalent in genetic programming \cite{overviewGP}.

\section{Distributed Evolutionary Computing}

The fact that evolutionary computing manipulates a population of independent solutions actually makes it well suited for parallel computation architectures \cite{cantupaz1998spg}. The motivation for using parallel or distributed evolutionary algorithms is twofold. First, improving the speed of evolutionary processes by conducting concurrent evaluations of individuals in a population. Second, improving the problem-solving process by overcoming difficulties that face traditional evolutionary algorithms, such as maintaining diversity to avoid premature convergence \cite{muhlenbein1991eta, stender1993pga}. There are several variants of distributed evolutionary computing, leading some to propose a taxonomy for their classification \cite{nowostawski1999pga}, with there being two main forms of models \cite{cantupaz1998spg, stender1993pga}:

\begin{itemize}
\item multiple-population/coarse-grained migration/island
\item single-population/fine-grained diffusion/neighbourhood
\end{itemize}

In the coarse-grained \emph{island} models \cite{lin1994cgp, cantupaz1998spg}, evolution occurs in multiple parallel sub-populations (islands), each running a local evolutionary algorithm, evolving independently with occasional \emph{migrations} of highly fit individuals among sub-populations. The core parameters for the evolutionary algorithm of the island-models are as follows \cite{lin1994cgp}:

\begin{itemize}
\item number of the sub-populations: 2, 3, 4, more
\item sub-population homogeneity
\begin{itemize}
\item size, crossover rate, mutation rate, migration interval
\end{itemize}
\item topology of connectivity: ring, star, fully-connect, random
\item static or dynamic connectivity
\item migration mechanisms: 
\begin{itemize}
\item isloated/synchronous/asynchronous
\item how often migrations occur 
\item which individuals migrate 
\end{itemize}
\end{itemize}

\tfigure{scale=0.75}{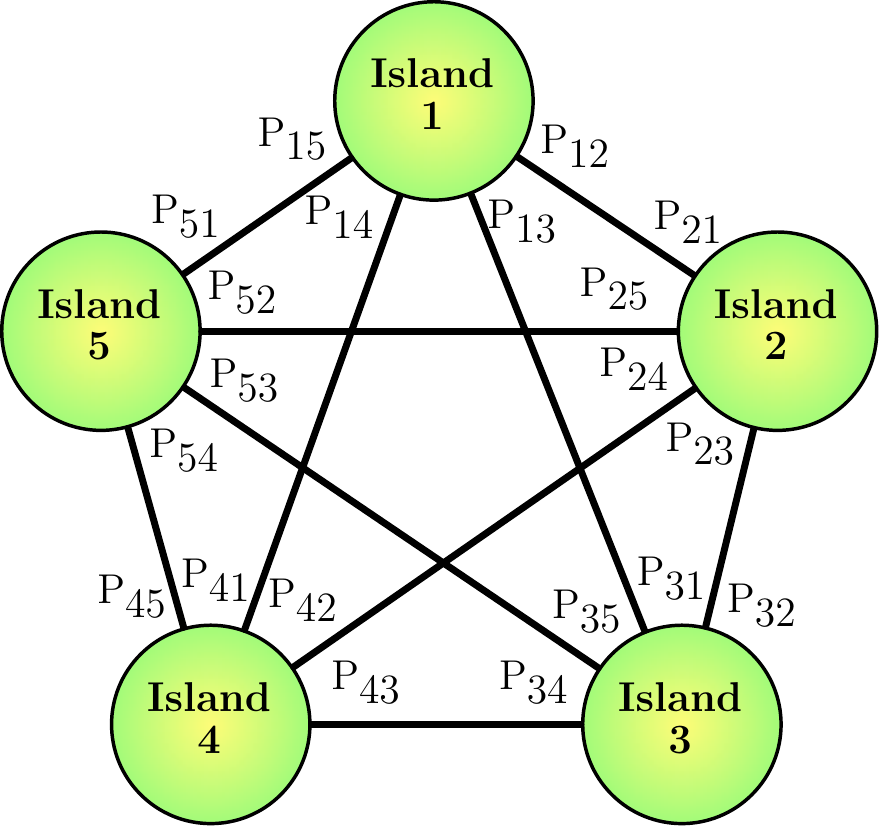}{graffle}{Island-Model of Distributed Evolutionary Computing}{\cite{lin1994cgp, cantupaz1998spg}: There \getCap{im1} This \getCap{im2}}{-2mm}{}{}{-4mm}

Fine-grained \emph{diffusion} models \cite{manderick1989fgp, stender1993pga} assign one individual per processor. A local neighbourhood topology is assumed, and individuals are allowed to mate only within their neighbourhood, called a \emph{deme}. The demes overlap by an amount that depends on their shape and size, and in this way create an implicit migration mechanism. Each processor runs an identical evolutionary algorithm which selects parents from the local neighbourhood, produces an offspring, and decides whether to replace the current individual with an offspring. However, even with the advent of multi-processor computers, and more recently multi-core processors, which provide the ability to execute multiple threads simultaneously \cite{newsArticle3}, this approach would still prove impractical in supporting the number of agents necessary to create a Digital Ecosystem. Therefore, we shall further consider the \emph{island} models.

An example island-model \cite{lin1994cgp, cantupaz1998spg} is visualised in Figure \ref{islandModel}, in which there \setCap{are different probabilities of going from island \Circ{1} to island \Circ{2}, as there is of going from island \Circ{2} to island \Circ{1}.}{im1} This allows maximum flexibility for the migration process, and \setCap{mirrors the naturally inspired quality that although two populations have the same physical separation, it may be easier to migrate in one direction than the other, i.e. fish migration is easier downstream than upstream.}{im2} The migration of the \emph{island} models is like the notion of migration in nature, being similar to the metapopulation models of theoretical ecology \cite{levins1969sda}. This model has also been used successfully in the determination of investment strategies in the commercial sector, in a product known as the Galapagos toolkit \cite{galapagos1, galapagos2}. However, all the \emph{islands} in this approach work on exactly the same problem, which makes it less analogous to biological ecosystems in which different locations can be environmentally different \cite{begon96}. We will take advantage of this property later when defining the \acl{EOA} of Digital Ecosystems.

\section{The Digital Ecosystem}
We are concerned with the digital counterpart of biological ecosystems. However, the term \emph{digital ecosystem} has been used to describe a variety of concepts, which it now makes sense to review. Some of these refer to the existing networking infrastructure of the internet \cite{debook2, fiorina, XIMBIOTIX}, while several companies offer a \emph{digital ecosystem} service or solution, which involves enabling customers to use existing e-business solutions \cite{accenture, syntel, xewow}. The term is also being increasingly linked, yet undefined, to the future developments of \ac{ICT} adoption for e-business and e-commerce, to create so called \emph{business ecosystems} \cite{iansiti2004kan, nachira, papazoglou2001aot}. However, perhaps the most frequent references to \emph{digital ecosystems} arise in Artificial Life research, where they are created primarily to investigate aspects of biological and other complex systems \cite{sorakugun1995eas, grand1998ces, deAI1}. The extent to which these disparate systems resemble biological ecosystems varies, and frequently the word \emph{ecosystem} is merely used for branding purposes without any inherent ecological properties.

We consider Digital Ecosystems \cite{javaOne, eveNet, eveSim} to be software systems that exploit the properties of biological ecosystems, which are robust, scalable, and self-organising \cite{Levin}. So, Digital Ecosystems provide a two-level optimisation scheme inspired by natural ecosystems, in which a decentralised peer-to-peer network forms an underlying tier of distributed agents. These agents then feed a second optimisation level based on an evolutionary algorithm that operates locally on single habitats (peers), aiming to find solutions that satisfy locally relevant constraints. The local search is sped up through this twofold process, providing better local optima as the distributed optimisation provides prior sampling of the search space by making use of computations already performed in other peers with similar constraints \cite{javaOne}. The agents consist of an \emph{executable component} and an \emph{ontological description} \cite{wooldridge}. So, the Digital Ecosystem can be considered a \ac{MAS} \cite{wooldridge} which uses \emph{distributed evolutionary computing} \cite{cantupaz1998spg, stender1993pga} to combine suitable agents in order to meet user requests for applications.

\tfigure{scale=1.0}{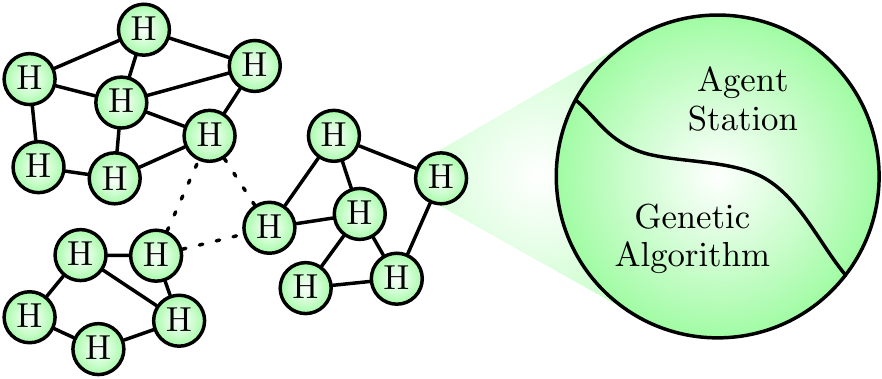}{graffle}{Digital Ecosystem}{Optimisation architecture in which agents (representing services) travel along the peer-to-peer connections; in every node (habitat) local optimisation is performed through an evolutionary algorithm, where the search space is determined by the agents present at the node.}{-7mm}{}{}{-4mm}

The landscape, in energy-centric biological ecosystems, defines the connectivity between habitats \cite{begon96}.  Connectivity of nodes in the digital world is generally not defined by geography or spatial proximity, but by information or semantic proximity. For example, connectivity in a peer-to-peer network is based primarily on bandwidth and information content, and not geography. The island-models of \acl{DEC} use an information-centric model for the connectivity of nodes (\emph{islands}) \cite{lin1994cgp}. However, because it is generally defined for one-time use (to evolve a solution to one problem and then stop) it usually has a fixed connectivity between the nodes, and therefore a fixed topology \cite{cantupaz1998spg}. So, supporting evolution in the Digital Ecosystem, with a multi-objective \emph{selection pressure} (fitness landscape \cite{wright1932} with many peaks), requires a re-configurable network topology, such that habitat connectivity can be dynamically adapted based on the observed migration paths of the agents between the users within the habitat network. Based on the island-models of \acl{DEC} \cite{lin1994cgp}, each connection between the habitats is bi-directional and there is a probability associated with moving in either direction across the connection, with the connection probabilities affecting the rate of migration of the agents. However, additionally, the connection probabilities will be updated by the success or failure of agent migration using the concept of Hebbian learning \cite{hebb}: the habitats which do not successfully exchange agents will become less strongly connected, and the habitats which do successfully exchange agents will achieve stronger connections. This leads to a topology that adapts over time, resulting in a network that supports and resembles the connectivity of the user base. If we consider a \emph{business ecosystem}, network of \aclp{SME}, as an example user base; such business networks are typically small-world networks \cite{white2002nst, antionella}. They \setCap{many strongly connected clusters (communities), called \emph{sub-networks} (quasi-complete graphs), with a few connections between these clusters (communities) \cite{swn1}. Graphs with this topology have a very high clustering coefficient and small characteristic path lengths \cite{swn1}.}{archComTop} So, the Digital Ecosystem will take on a topology similar to that of the user base.

The novelty of our approach comes from the evolving populations being created in response to \emph{similar} requests. So whereas in the island-models of \acl{DEC} there are multiple evolving populations in response to one request \cite{lin1994cgp}, here there are multiple evolving populations in response to \emph{similar} requests. In our Digital Ecosystems different \setCap{requests are evaluated on separate \emph{islands} (populations), and so adaptation is accelerated by the sharing of solutions between evolving populations (islands), because they are working to solve similar requests (problems).}{similarCap}

The users \setCap{will formulate queries to the Digital Ecosystem by creating a request as a \emph{semantic description}, like those being used and developed in \acp{SOA} \cite{SOAsemantic}, specifying an application they desire and submitting it to their local peer (habitat).}{picUser} This description defines a metric for evaluating the \emph{fitness} of a composition of agents, as a distance function between the \emph{semantic description} of the request and the agents' \emph{ontological descriptions}. \setCap{A population is then instantiated in the user's habitat in response to the user's request, seeded from the agents available at their habitat.}{picUserReq} This allows the evolutionary optimisation to be accelerated in the following three ways: first, the habitat network provides a subset of the agents available globally, which is localised to the specific user it represents; second, making use of agent-sequences previously evolved in response to the user's earlier requests; and third, taking advantage of relevant agent-sequences evolved elsewhere in response to similar requests by other users. The population then proceeds to evolve the optimal agent-sequence(s) that fulfils the user request, and as the agents are the base unit for evolution, it searches the available agent-sequence combination space. For an evolved agent-sequence that is executed (instantiated) by the user, it then migrates to other peers (habitats) becoming hosted where it is useful, to combine with other agents in other populations to assist in responding to other user requests for applications.

\section{Simulation and Results}

An important measure for determining the success of the Digital Ecosystem is its relative performance to a traditional \ac{SOA} based system. So, we simulated a simple SOA with a distributed UDDI service registry \cite{papazoglou2003soc}, with redirects (links to other nodes) at each node for service descriptions not stored locally, and without caching (analogous to the Digital Ecosystem). We then compared it against a typical simulation run of the Digital Ecosystem. The time available to the SOA based system for fulfilling a request from the distributed registry was limited to the time the Digital Ecosystem required to respond to the same user request, and worked by searching for the optimal services based on the optimal segmentation of the request, because an exhaustive combinatorial search would have been impractical. We simulated the Digital Ecosystem, following the \acl{EOA} from the previous section. Throughout the simulations we assumed a hundred users, which meant that at any time the number of users joining the network equalled those leaving. The habitats of the users were randomly connected at the start, to simulate the users going online for the first time. The users then produced agents (services) and requests for business applications. Initially, the users each deployed five agents to their habitats, for migration (distribution) to any habitats connected to theirs (i.e. their community). Users were simulated to deploy a new agent after the submission of three requests for business applications, and were chosen at random to submit their requests. 

\setCap{A simulated user request consisted of an abstract \emph{semantic description}, as a list of sets of numeric tuples to represent the properties of a desired business application}{semanticRequest}. The use of the \emph{numeric tuples} made it comparable to the \emph{semantic descriptions} of the services represented by the agents; while the \emph{list of sets} (two level hierarchy) and a much longer length provided sufficient complexity to support the sophistication of business applications.

In the Digital Ecosystem user requests were handled by the habitats instantiating evolving populations, which used evolutionary computing to find the optimal solution(s), agent-sequence(s). It was assumed that the users made their requests for business applications \emph{accurately}, and always used the response (agent-sequence) provided. Populations of agents, $[A_1, A_1, A_2, ...]$, were evolved to solve user requests, seeded with agents and agent-sequences from the \emph{agent-pool} of the habitats in which they were instantiated. A dynamic population size was used to ensure exploration of the available combinatorial search space, which increased with the average length of the population's agent-sequences. The optimal combination of agents (agent-sequence) was evolved to the user request, $R$, by an artificial \emph{selection pressure} created by a \emph{fitness function} generated from the user request, $R$. An individual (agent-sequence) of the population consisted of a set of attributes, ${a_1, a_2, ...}$, and a user request essentially consisted of a set of required attributes varied according to a Gaussian distribution, ${r_1, r_2, ...}$. So, the \emph{fitness function} for evaluating an individual agent-sequence, $A$, relative to a user request, $R$, was,
\vspace{6mm}
\begin{equation}
fitness(A,R) = \frac{1}{1 + \sum_{r \in R}{|r-a|}},
\label{ff}
\vspace{6mm}
\end{equation}
where $a$ is the member of $A$ such that the difference to the required attribute $r$ was minimised. Equation \ref{ff} was used to assign \emph{fitness} values between 0.0 and 1.0 to each individual of the current generation of the population, directly affecting their ability to replicate into the next generation. The evolutionary computing process was encoded with a low mutation rate, a fixed selection pressure and a non-trapping fitness function (i.e. did not get trapped at local optima). The type of selection used \emph{fitness-proportional} and \emph{non-elitist}. \emph{Fitness-proportional meaning that the \emph{fitter} the individual the higher its probability of} surviving to the next generation \cite{blickle1996css}. \emph{Non-elitist} meaning that the best individual from one generation was not guaranteed to survive to the next generation; it had a high probability of surviving into the next generation, but it was not guaranteed as it might have been mutated, \cite{eiben2003iec}. \emph{Crossover} (recombination) was then applied to a randomly chosen 10\% of the surviving population, a \emph{one-point crossover}, by aligning two parent individuals and picking a random point along their length, and at that point exchanging their tails to create two offspring \cite{eiben2003iec}. \emph{Mutations} were then applied to a randomly chosen 10\% of the surviving population; \emph{point mutations} were randomly located, consisting of \emph{insertions} (an agent was inserted into an agent-sequence), \emph{replacements} (an agent was replaced in an agent-sequence), and \emph{deletions} (an agent was deleted from an agent-sequence) \cite{lawrence1989hsd}. The issue of bloat was controlled by augmenting the \emph{fitness function} with a \emph{parsimony pressure \cite{soule1998ecg} which biased the search} to shorter agent-sequences, evaluating longer than average length agent-sequences with a reduced \emph{fitness}, and thereby providing a dynamic control limit which adapted to the average length of the ever-changing evolving agent populations.

\tfigure{scale=1.0}{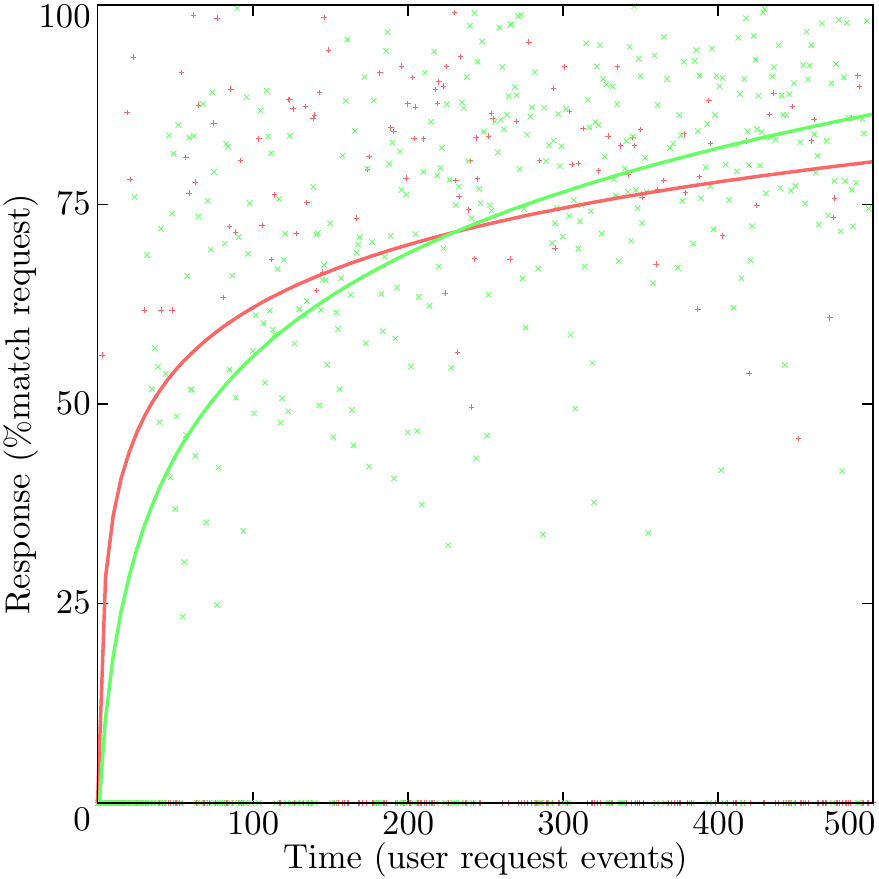}{pdf}{Graph of the performance of the Digital Ecosystem against a traditional \ac{SOA} based system}{\getCap{newCapGraph}}{-7mm}{}{}{-4mm}

In Figure \ref{newGraph} we graphed the percentage match to the user requests for typical runs, as determined by a distance function between the request and the service descriptions. \setCap{Both the Digital Ecosystem and the \ac{SOA} based system performed as expected, providing better responses to user requests as more services became available. The SOA reference system initially performed better than the Digital Ecosystem, but with the increasing number of services the Digital Ecosystem outperformed the reference system.}{newCapGraph} This was anticipated as the the Digital Ecosystem was expected to be more effective at larger scales \cite{javaOne}.

\section{Conclusion}

The experimental results indicate that under simulation conditions the Digital Ecosystem outperforms the comparison system based on a traditional \acl{SOA}. Service-oriented architectures promise to provide potentially huge numbers of services that programmers can combine via standardised interfaces, to create increasingly sophisticated and distributed applications \cite{SOApaper2}. The Digital Ecosystem extends this concept with the automatic combining of available and applicable services in a scalable architecture to meet user requests for applications. This is made possible by a fundamental paradigm shift, from a \emph{pull}-oriented approach to a \emph{push}-oriented approach. So, instead of the \emph{pull}-oriented approach of generating applications only upon request in \aclp{SOA} \cite{singh2005soc}, the Digital Ecosystem follows a \emph{push}-oriented approach of distributing and composing applications pre-emptively, as well as upon request. Although the use of \aclp{SOA} in the definition of Digital Ecosystems provides a predisposition to business \cite{krafzig2004ess}, it does not preclude other more general uses. The \acl{EOA} definition of Digital Ecosystems is intended to be inclusive and interoperable with other technologies, in the same way that the definition of \aclp{SOA} is with \emph{grid computing and other} technologies \cite{singh2005soc}. For example, habitats could be executed using a distributed processing arrangement, such as \emph{cloud computing} \cite{weiss2007cc}, which would be possible because the habitat network topology is information-centric (instead of location-centric).

\section*{Acknowledgment}
The authors would like to thank for their encouragement and suggestions: Paolo Dini of the \acl{LSE}, Thomas Heistracher and his group of the \acl{STU}, Jonathan Rowe of the \acl{UBHAM}, and Miguel Vidal of \acl{SUN}. The Digital Ecosystem model was constructed through interacting with these people and others.

\bibliographystyle{IEEEtran.bst}
\bibliography{/Users/g/Desktop/PhDthesis/references}

\end{document}

%% file: acronyms.tex
\acrodef{PCG}{Projected Conjugate Gradient} 
\acrodef{QP}{quadratic programming}
\acrodef{RBF}{Radial-Basis Function}
\acrodef{ABM}{Agent-Based Modelling}
\acrodef{AI}{Artificial Intelligence}
\acrodef{DAI}{Distributed Artificial Intelligence}
\acrodef{API}{Application Programming Interface}
\acrodef{ARF}{p14ARF human tumor-suppressor gene}
\acrodef{B2B}{business-to-business}
\acrodef{BDP}{Biological Design Pattern}
\acrodef{BGS}{Best Guess Solution}
\acrodef{BIC}{Biologically-Inspired Computing}
\acrodef{BML}{Business Modelling Language}
\acrodef{BPEL}{Business Process Execution Language}
\acrodef{BPMN}{Business Process Modelling Notation}
\acrodef{CAS}{Complex Adaptive Systems}
\acrodef{COBOL}{COmmon Business-Oriented Language}
\acrodef{DBE}{Digital Business Ecosystem}
\acrodef{DE}{Digital Ecosystem}
\acrodef{DEC}{distributed evolutionary computing}
\acrodef{DGA}{Distributed genetic algorithms}
\acrodef{DIS}{Distributed Intelligence System}
\acrodef{DNA}{Deoxyribose Nucleic Acid}
\acrodef{DOP}{DBE Open Protocol}
\acrodef{DSS}{Distributed Storage System}
\acrodef{EAP}{Evolving Agent Population}
\acrodef{ebXML}{e-business eXtensible Markup Language}
\acrodef{EC}{Evolutionary Computing}
\acrodef{ECJ}{Evolutionary Computing in Java}
\acrodef{EE}{Evolutionary Environment}
\acrodef{EFL}{Evolutionary Framework for Language}
\acrodef{FLE}{Framework for Language Ecosystems}
\acrodef{EOA}{Ecosystem-Oriented Architecture}
\acrodef{ESS}{evolutionary stable strategy}
\acrodef{EvE}{Evolutionary Environment}
\acrodef{ExE}{Execution Environment}
\acrodef{FCB}{Framework for Computational Biomimicry}
\acrodef{FFF}{Fitness Function Framework}
\acrodef{FL}{Fitness Landscape}
\acrodef{HWU}{Heriot-Watt University}
\acrodef{ICL}{Imperial College London}
\acrodef{ICT}{Information and Communications Technology}
\acrodef{INTEL}{Intel Ireland}
\acrodef{IPA}{International Phonetic Alphabet}
\acrodef{ISUFI}{Istituto Superiore Universitario di Formazione Interdisciplinare}
\acrodef{JDJ}{Java Developer's Journal}
\acrodef{KC}{Kolmogorov-Chaitin}
\acrodef{LAN}{local area network}
\acrodef{LSE}{London School of Economics and Political Science}
\acrodef{MAS}{Multi-Agent System}
\acrodef{MDL}{Minimum Description Length}
\acrodef{MDM2}{murine double minute 2}
\acrodef{MFT}{Mean Field Theory}
\acrodef{MoAS}{Mobile Agent System}
\acrodef{MOF}{Meta Object Facility}
\acrodef{MUH}{migration and usage history}
\acrodef{NIC}{Nature Inspired Computing}
\acrodef{NN}{Neural Network}
\acrodef{NoE}{Network of Excellence}
\acrodef{OMG}{Open Mac Grid}
\acrodef{OPAALS}{Open Philosophies for Associative Autopoietic Digital Ecosystems}
\acrodef{P2P}{peer-to-peer}
\acrodef{P53}{protein 53}
\acrodef{PDA}{Personal Digital Assistant}
\acrodef{QoS}{quality of service}
\acrodef{REST}{REpresentational State Transfer}
\acrodef{RNA}{Deoxyribose Nucleic Acid}
\acrodef{SAE}{Software Agent Ecosystem}
\acrodef{SBML}{Systems Biology Modelling Language}
\acrodef{SBVR}{Semantics of Business Vocabulary and Business Rules}
\acrodef{SDL}{Service Description Language}
\acrodef{SF}{Service Factory}
\acrodef{SIM}{Social Interaction Mechanism}
\acrodef{SM}{Service Manifest}
\acrodef{SME}{Small and Medium sized Enterprise}
\acrodef{SML}{Service Modelling Language}
\acrodef{SMO}{Sequential Minimal Optimisation}
\acrodef{SOA}{Service-Oriented Architecture}
\acrodef{SOAP}{Simple Object Access Protocol}
\acrodef{SOC}{Self-Organised Criticality}
\acrodef{SOLUTA}{SOLUTA.NET}
\acrodef{SOM}{Self-Organising Map}
\acrodef{SSL}{Semantic Service Language}
\acrodef{STU}{Salzburg Technical University}
\acrodef{SUN}{Sun Microsystems}
\acrodef{SVM}{Support Vector Machine}
\acrodef{TM}{Turing Machine}
\acrodef{UBHAM}{University of Birmingham}
\acrodef{UDDI}{Universal Description Discovery and Integration}
\acrodef{UML}{Unified Modelling Language}
\acrodef{URI}{Uniform Resource Identifier}
\acrodef{UTM}{Universal Turing Machine}
\acrodef{VLP}{variable length population}
\acrodef{VLS}{variable length sequences}
\acrodef{vls}{variable length sequence}
\acrodef{WP}{Work-Package}
\acrodef{WSDL}{Web Services Definition Language}
\acrodef{XMI}{XML Metadata Interchange}
\acrodef{XML}{eXtensible Markup Language}
\acrodef{MD5}{Message-Digest algorithm 5}
\acrodef{GA}{genetic algorithm}
\acrodef{GP}{genetic programming}
\acrodef{MASON}{Multi-Agent Simulator Of Neighbourhoods}
\acrodef{Repast}{Recursive Porous Agent Simulation Toolkit}
\acrodef{JCLEC}{Java Computing Library for Evolutionary Computing}
\acrodef{OWL-S}{Web Ontology Language - Service}
\acrodef{EGT}{Evolutionary Game Theory}
\acrodef{RBF}{Radial Basis Functions}
\acrodef{SWS}{Semantic Web Services}
\acrodef{HDD}{Hard Disk Drive}
\acrodef{SSD}{Solid-State Drive}

%% file: captions.tex
\acrodef{im1}{are different probabilities of going from island \Circ{1} to island \Circ{2}, as there is of going from island \Circ{2} to island \Circ{1}.}
\acrodef{im2}{mirrors the naturally inspired quality that although two populations have the same physical separation, it may be easier to migrate in one direction than the other, i.e. fish migration is easier downstream than upstream.}
\acrodef{archComTop}{many strongly connected clusters (communities), called {sub-networks} (quasi-complete graphs), with a few connections between these clusters (communities) \cite{swn1}. Graphs with this topology have a very high clustering coefficient and small characteristic path lengths \cite{swn1}.}
\acrodef{similarCap}{requests are evaluated on separate {islands} (populations), and so adaptation is accelerated by the sharing of solutions between evolving populations (islands), because they are working to solve similar requests (problems).}
\acrodef{picUser}{will formulate queries to the Digital Ecosystem by creating a request as a {semantic description}, like those being used and developed in \acp{SOA} \cite{SOAsemantic}, specifying an application they desire and submitting it to their local peer (habitat).}
\acrodef{picUserReq}{A population is then instantiated in the user's habitat in response to the user's request, seeded from the agents available at their habitat.}
\acrodef{semanticRequest}{A simulated user request consisted of an abstract {semantic description}, as a list of sets of numeric tuples to represent the properties of a desired business application}
\acrodef{newCapGraph}{Both the Digital Ecosystem and the \ac{SOA} based system performed as expected, providing better responses to user requests as more services became available. The SOA reference system initially performed better than the Digital Ecosystem, but with the increasing number of services the Digital Ecosystem outperformed the reference system.}